\documentclass[10pt,twocolumn,letterpaper]{article}

\usepackage{iccv}              
\usepackage{pifont}
\usepackage{bbding} 
\usepackage{threeparttable}
\usepackage{multirow}
\usepackage{indentfirst}
\usepackage{cuted}
\usepackage{microtype}
\usepackage{capt-of}
\usepackage{booktabs} 
\usepackage{enumitem}
\usepackage{amsmath}
\usepackage[ruled,vlined]{algorithm2e}
\usepackage{algorithmicx}
\usepackage{algpseudocode}
\algdef{SE}[DOWHILE]{Do}{doWhile}{\algorithmicdo}[1]{\algorithmicwhile\ #1}%

\newcommand{\yl}[1]{{\color{black}#1}}
\newcommand{\cg}[1]{{\color{black}#1}}
\newcommand{\mj}[1]{{\color{black}#1}}
\newcommand{\cmark}{\ding{52}} 
\newcommand{\xmark}{\ding{56}} 
\definecolor{iccvblue}{rgb}{0.21,0.49,0.74}
\usepackage[pagebackref,breaklinks,colorlinks,allcolors=iccvblue]{hyperref}

\def\paperID{4201} 
\def\confName{ICCV}
\def\confYear{2025}

\title{RESCUE: Crowd Evacuation Simulation via Controlling SDM-United Characters}

\author{
    Xiaolin Liu\textsuperscript{1,\dag}, Tianyi Zhou\textsuperscript{1,\dag}, Hongbo Kang\textsuperscript{1}, Jian Ma\textsuperscript{1}, Ziwen Wang\textsuperscript{1}, \\
    Jing Huang\textsuperscript{1}, Wenguo Weng\textsuperscript{2}, Yu-Kun Lai\textsuperscript{3}, Kun Li\textsuperscript{1,*} \\
    \textsuperscript{1}Tianjin University, \textsuperscript{2}Tsinghua University, \textsuperscript{3}Cardiff University \\
    {\tt\small \{liuxiaolin, tianyizhou, hbkang, jianma, 3021244417, hj00, lik\}@tju.edu.cn}, \\
    {\tt\small wgweng@tsinghua.edu.cn, Yukun.Lai@cs.cardiff.ac.uk} \\
    \textsuperscript{\dag}Equal contribution
    \textsuperscript{*}Corresponding author
}

\begin{document}

\maketitle

\begin{strip}\centering
\vspace{-5.0em}
\includegraphics[width=\textwidth]{imgs/Top_v2.pdf}
\captionof{figure}{Our SDM-United Framework enables \cg{online} simulation of personalized, physically realistic, and 3D-adaptive multi-agent evacuation scenarios.
\label{fig:fig_teaser}
}
\vspace{-1.0em}
\end{strip}

\begin{abstract}

\vspace{-1.0em}

Crowd evacuation simulation is critical for enhancing public safety, and demanded for realistic virtual environments. Current mainstream evacuation models overlook the complex human behaviors that occur during evacuation, such as pedestrian collisions, interpersonal interactions, and variations in behavior influenced by terrain types or individual body shapes. This results in the failure to accurately simulate the escape of people in the real world. In this paper, aligned with the sensory-decision-motor (SDM) flow of the human brain, we propose a real-time 3D crowd evacuation simulation framework that integrates a 3D-adaptive SFM (Social Force Model) Decision Mechanism and a Personalized Gait Control Motor. This framework allows multiple agents to move in parallel and is suitable for various scenarios, with dynamic crowd awareness. Additionally, we introduce Part-level Force Visualization to assist in evacuation analysis. Experimental results demonstrate that our framework supports dynamic trajectory planning and personalized behavior for each agent throughout the evacuation process, and is compatible with uneven terrain. Visually, our method generates evacuation results that are more realistic and plausible, providing enhanced insights for crowd simulation. The code is available at http://cic.tju.edu.cn/faculty/likun/projects/RESCUE.

\end{abstract}    
\vspace{-2em}
\section{Introduction}
\label{sec:intro}

\cg{Evacuation simulations serve as valuable tools for assessing the likelihood of crowding and trampling incidents, estimating evacuation times, and 
\yl{supporting}
virtual reality escape training. However, no existing methods can simulate personalized and proxemics-aware 3D movements of hundreds of people online.} This paper aims to develop an adaptive multi-agent 3D evacuation simulation framework. As shown in Figure~\ref{fig:fig_teaser}, our framework, called RESCUE (cRowd Evacuation Simulation via Controlling SDM-United charactErs), is capable of achieving online physically realistic simulations performing avoidance and personalized gait. 

\begin{table*}[h]
\centering
\footnotesize
\vspace{-1.5em}
\caption{Comparison with Related Methods}
\vspace{-0.5em}
\renewcommand{\arraystretch}{1}
\begin{tabular}{c|p{2.6cm}|p{2.6cm}|p{2.6cm}|c}
\hline
Related Work & Traditional Crowd Simulation \cite{DING2024128448,NGUYEN201344,Chen01012008,doi:10.1061/(ASCE)CP.1943-5487.0000532} & Path-Guided Motion Generation \cite{wan2023tlcontrol,xie2023omnicontrol,pinyoanuntapong2024controlmm} & Path-Guided Character Control \cite{rempe2023trace,wang2024pacer+, tessler2024maskedmimic} & Our Method \\ \hline
Online Path Planning & \multicolumn{1}{c|}{\textit{\textbf{Partial}}} & \multicolumn{1}{c|}{\xmark} & \multicolumn{1}{c|}{\xmark} & \multicolumn{1}{c}{\cmark} \\ \hline
3D Human Models & \multicolumn{1}{c|}{\xmark} & \multicolumn{1}{c|}{\cmark} & \multicolumn{1}{c|}{\cmark} & \multicolumn{1}{c}{\cmark} \\ \hline
Personalized Gait & \multicolumn{1}{c|}{\xmark} & \multicolumn{1}{c|}{\cmark} & \multicolumn{1}{c|}{\xmark} & \multicolumn{1}{c}{\cmark} \\ \hline
Collision Visualization & \multicolumn{1}{c|}{\xmark} & \multicolumn{1}{c|}{\xmark} & \multicolumn{1}{c|}{\xmark} & \multicolumn{1}{c}{\cmark} \\ \hline
Uneven Terrains & \multicolumn{1}{c|}{\xmark} & \multicolumn{1}{c|}{\xmark} & \multicolumn{1}{c|}{\cmark} & \multicolumn{1}{c}{\cmark} \\ \hline
\end{tabular}
\vspace{-1em}

\label{tab:comparison}
\vspace{-1.0em}
\end{table*}

Traditional crowd simulation methods \cite{DING2024128448,NGUYEN201344,Chen01012008} have explored various strategies in different scenarios. However, due to their simplified representations, these methods fail to integrate 3D motion with realistic behavior, resulting in physically implausible actions. Most crowd simulation methods cannot make decisions based on scene changes and are not designed for crowd evacuation simulation. \cg{Based on physics engines \cite{makoviychuk2021isaac}, character control \cite{tessler2023calm,2022-TOG-ASE,10.1145/1778765.1781156,rempe2023trace,tessler2024maskedmimic} achieves autonomous 3D motion.} However, these
methods face the lack of personalization motions and may result in  falls and collisions in densely populated scenarios. \cg{Some \yl{motion generation methods} \cite{Guo_2022_CVPR,shafir2023human,shi2023controllable,xie2023omnicontrol} based on diffusion models can generate diverse 3D motions}, but they still face challenges in controllability and physical realism. Neither of these methods can directly produce realistic, personalized, and online evacuation motions. The detailed capabilities of the methods are summarized in Table~\ref{tab:comparison}.

\cg{The limitations of the existing methods are mainly due to the significant behavioral and environmental complexities inherent in evacuation scenarios: 1) \textit{3D Proxemics-aware Ability}: Crowded pathways and frequent physical interactions require the use of online 3D movement decision to enable dynamic strategy adjustments, such as avoidance of collisions and maintaining stability; 2) \textit{Personalized Gait}: Individuals with different attributes, such as older or disabled persons, exhibit varying behaviors in identical situations, adapting to the scene with different speed and performance. }

\cg{Neuroscientific studies \cite{khilkevich2024brain,siegel2015cortical,miller2001integrative} demonstrate that humans employ a Sensory-Decision-Motor (SDM) loop~\cite{PLATT2002141}, integrating environmental cues through distributed neural networks to evaluate and dynamically adjust behavior in complex environments. In this paper, we propose \yl{an} online \textbf{SDM-United 3D Evacuation Simulation Framework} that integrates personalized decision-making with character control in a physics engine. This allows each agent to perceive its surroundings and make online adjustments.} To handle congestion, we propose a \textit{3D-Adaptive Social Force Model} for decision-making and obstacle avoidance in 3D environments. In particular, we introduce an evasive force that adapts to dense scenarios to avoid congestion.

To achieve behavioral diversity, we introduce a \textit{Proxemics-aware Personalization Method} with a personalized gait controller and an optimized SFM to generate individualized behaviors for agents based on attributes such as ages and physical conditions. Additionally, we introduce \textit{Part-level Force Visualization}, offering unprecedented insights into contact forces. Our experiments demonstrate that the proposed framework effectively adapts to uneven terrain, maintains balance during collisions and crowd congestion, and generates personalized behaviors. 
Although our experiments have focused on evacuation scenarios, the framework readily generalizes to a wide variety of multi-agent environments, such as autonomous driving simulations in which dynamic vehicles and diverse pedestrians coexist. 
Our contributions can be summarized as follows:
\begin{itemize}
\item We propose the first online SDM-United 3D Evacuation Simulation Framework, with online decision-making and physics-based motion generation. This proxemics aware framework can be seamlessly extended to arbitrary dynamic environments and supports real time parallel simulation of hundreds of agents.
\item We propose a 3D-adaptive Social Force Model, designing optimized forces and personalized coefficients to enable accurate decision-making in 3D environments. This approach ensures correct guidance for the speed of agents.
\item We propose a Personalized Gait Controller, enabling agents to generate personalized motions tailored to attributes such as age and physical condition. Additionally, a Part-level Force Visualization is designed for analysis.
\item We validate the framework in various evacuation scenarios, demonstrating its ability to simulate diverse evacuation events, including the effects of crowd density, corridor width and terrain influence trampling incidents, as well as crowd behaviors like collisions and falls.
\end{itemize}
\begin{figure*}[!ht]
\vspace{-2em}
\centering
\includegraphics[scale=0.07]{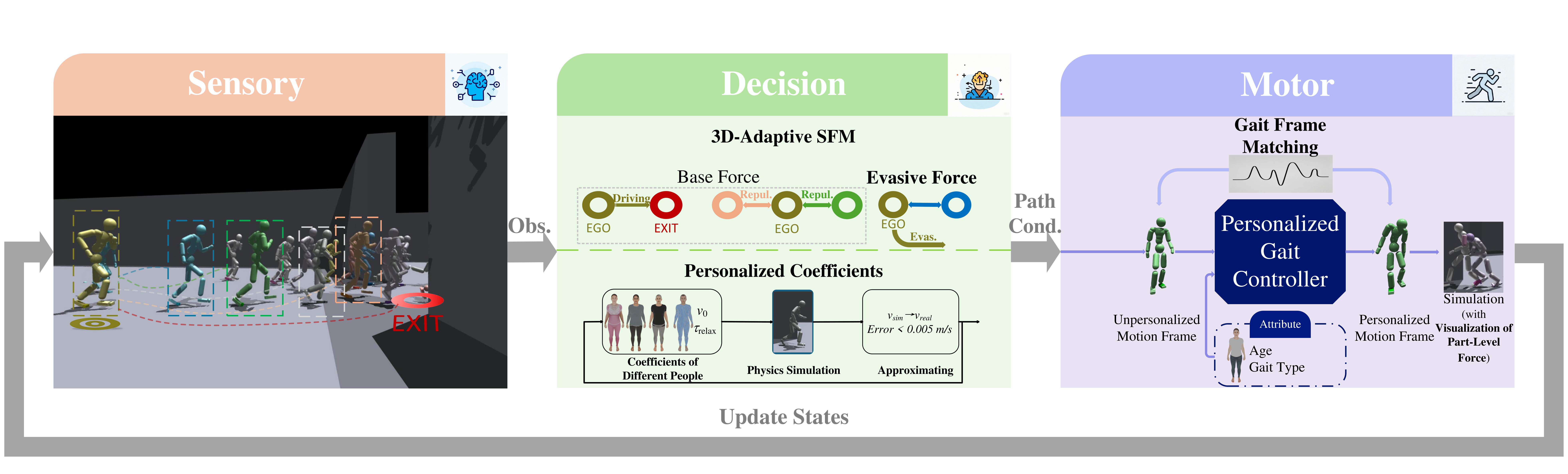}

		\vspace{-1em}
        
\caption{The detailed architecture of RESCUE. Our framework simulates the neuroscience-validated \cg{paradigm} (Sensory-Decision-Motor) to achieve a realistic and personalized crowd evacuation simulation. Specifically, our framework includes: (a) 
Sensory: each agent senses observations, including the fully observable self-state, the partially observable other-state, and the environment state; (b) Decision: each agent uses Social Force Model with personalized SFM coefficients to decide its speed in the next timestep, which is then used to compute path condition; and (c) Motor: the path condition and personal attribute of each agent are conditioned to simulate locomotion with a personalized gait controller. The simulation then updates the states of all agents, allowing them to be sensed by the sensory module.}
		\label{fig-framework}
    \vspace{-1.5em}
    
	\end{figure*}

\vspace{-0.5em}

\section{Related Work}
\label{sec:formatting}

\cg{
\subsection{Crowd Simulation}}

Crowd simulation is a tool for analyzing and optimizing crowd dynamics. The aim is to capture key phenomena such as bottlenecks and panic-induced behaviors. Macroscopic models describe collective crowd movement using fluid dynamics \cite{hamacher2001mathematical,cristiani2014multiscale,5481133,GEORGOUDAS2010285}, while providing detailed insights through approaches. Social Force Models (SFM) \cite{helbing1995social, helbing2000simulating,shuaib2017incorporating} are among the most widely used methods, treating pedestrians as Newtonian particles influenced by the forces driving acceleration. Additional forces such as panic, sliding friction, and body compression have been introduced \cite{DING2024128448,Zainuddin13012010}. However, these refinements often target specific scenarios. With advances in computational power, Agent-Based Models (ABM) have gained popularity \cite{Chen01012008,doi:10.1061/(ASCE)CP.1943-5487.0000532}, simulating individuals as autonomous agents with unique decision-making capabilities. Although the above-mentioned methods attempt to simulate real-world escape, they rely on simplified 2D representations, where force feedback can only use unrealistic surrogate computations, leading to distorted speed outputs. \cg{Although several attempts at physically plausible 3D simulations have emerged \cite{10.1145/3424636.3426894, ye2024crowd, charalambous2023greil, gomez2024resolving}, these approaches still fall short in integrating personalized motion control and decision-making mechanisms. Most methods rely on generic movement patterns that overlook individual variations and fail to construct comprehensive strategies for handling crowd dynamics and escape scenarios.}


\subsection{Character Control}
The practicable controllers for physics-based character simulation remain a challenge in animation. Initial methods predominantly rely on carefully designed control structures, such as finite state machines and trajectory optimization, to achieve specific motions. These approaches demonstrate the feasibility of simulating a wide range of behaviors, but their reliance on task-specific engineering makes them inflexible and time consuming to adapt to new tasks \cite{10.1145/1778765.1781156}. Recent advances in deep reinforcement learning (DRL) greatly expand the capabilities of physics-based characters, enabling them to perform diverse tasks, such as walking and running \cite{10.1145/3072959.3073602, xie2020allstepscurriculumdrivenlearningstepping}, and recovering from disturbances \cite{10.1145/3623264.3624448}. Designing effective reward functions can produce natural motions, but remains challenging. To address this, motion imitation methods \cite{Truong_2024, peng2021amp} gain popularity, where controllers learn from reference data, either through direct tracking or adversarial learning, to produce lifelike behaviors. Furthermore, hierarchical control frameworks are proposed to combine low-level imitation with high-level motion planning, enabling characters to execute complex sequences of actions \cite{rempe2023trace, tessler2024maskedmimic}. However, existing methods struggle to generate realistic behaviors in densely populated scenes, lack online decision-making capabilities, and are insufficient in generating diverse and personalized motions.

\subsection{Motion Generation}

Motion generation aims to synthesize realistic and varied human motions. Recent advances in diffusion models have significantly improved the diversity and quality of motion synthesis \cite{shafir2023human,shi2023controllable}. Key research directions include text-driven motion generation, stylized motion synthesis \cite{zhong2025smoodi,chen2024taming}, scene-aware motion generation \cite{wang2021scene,zhao2023synthesizing,pan2024synthesizing}, and human-object interaction modeling \cite{xu2023interdiff}. Recent efforts \cite{yuan2023physdiff,shi2024interactivecharactercontrolautoregressive} have also integrated physical constraints to enhance the realism of generated motions. Trajectory-controlled \cite{wan2023tlcontrol,xie2023omnicontrol,pinyoanuntapong2024controlmm} and multi-person motion synthesis \cite{DBLP:journals/corr/abs-2311-15864} have been explored. However, challenges remain in generating long-duration, physically plausible motions, particularly in multiperson, contact-rich scenarios. In addition, these methods are usually time consuming and not suitable for online applications.

In conclusion, all \yl{existing} methods fail to accurately simulate 3D multi-agent with personalized behaviors in densely populated escape scenarios. Therefore, we propose the first SDM-unified evacuation simulation framework to generate realistic, personalized, and online 3D evacuation motions.

\vspace{-0.5em}

\section{Method}
\subsection{Overview}
\cg{
The structure of our framework is shown in Figure \ref{fig-framework}. Our goal is to simulate realistic evacuation scenarios with diverse individuals. Given a scene mesh, \mj{an} exit location and initial positions, \mj{the framework} generates a personalized evacuation animation. \mj{Notably, it} is suitable for evacuation tasks by incorporating human-like paradigm, while ensuring personalized decisions and motions.

In Section \ref{Framework}, we introduce our SDM-United Framework, comprising the Sensory, Decision, and Motor modules. The Sensory module provides self-awareness, partial awareness of others, and exit perception. Our 3D-adaptive SFM Mechanism (Section \ref{Decision}) enhances decision-making
and obstacle avoidance in 3D environments. Proxemics-\yl{aware} Personalization is achieved through both the Decision and Motor modules: the Decision module, with Personalized Optimization for SDM Coefficients (Section \ref{SDM Coefficients}), computes speeds tailored to individual attributes (Section \ref{SDM Coefficients}), while the Motor module, with the Personalized Gait Controller (Section~\ref{Motor}), generates locomotion with personalized gaits. We also introduce the Visualization of Part-Level Forces (Section~\ref{Visualization}) to aid analysis.

\subsection{SDM-United Framework} \label{Framework}
The human brain follows a workflow of sensory, decision, and motor processes \cite{PLATT2002141}, and hence we develop our framework. Each person in the evacuation process is abstracted as an agent that operates in an online loop, where the sensory module gathers observations, the decision module computes speeds, and the motor module executes motions, with updated states feeding back into the sensory module for continuous processing and making them available for next cycle.

\textbf{Sensory.} Each agent perceives its self-state $\boldsymbol{s}^{self}$, the partially observable other-state $\boldsymbol{s}^{other}$, and the environment state $\boldsymbol{s}^{envir}$. At time $t$, $\boldsymbol{s}^{self}_{i,t}$ includes the agent's position, speed, and its humanoid state $\boldsymbol{h}_{i,t}$. The other-state $\boldsymbol{s}^{other}_{i,t}$ contains other agents' root positions relative to the agent, and the environment state $\boldsymbol{s}^{envir}_{i,t}$. The form of $\boldsymbol{h}_{i,t}$ and $\boldsymbol{s}^{envir}_{i,t}$ follows \cite{rempe2023trace}.

\textbf{Decision.} The Social Force Model (SFM) \cite{helbing1995social,helbing2000simulating} models pedestrian behavior, including panic and crowd dynamics. We propose a 3D-adaptive SFM Decision Mechanism to compute the desired speed $\tilde{\mathbf{v}}_{i,t+1}$. \mj{The key} points of the escape path are found \mj{via} A* search \cite{hart1968formal}, and we use the SFM to drive each agent toward the key points while avoiding collisions, resulting in the computation of the agent's speed $\tilde{\mathbf{v}}_{i,t+1}$. SFM coefficients are optimized based on individual attributes to enable personalized decisions. 

\textbf{Motor.} We use Pacer \cite{rempe2023trace} for path-following, which computes the motion $\boldsymbol{a}_{i,t}$ based on the speed $\tilde{\mathbf{v}}_{i,t+1}$. This motion is \textit{non-personalized}, so a Personality Gait Controller assigns a personalized gait based on attributes. The resulting motion is simulated in the physics engine and then sensed in by each agent for continuous processing.}
\subsection{3D-adaptive SFM Decision Mechanism} \label{Decision}
In 3D simulations, interaction dynamics differs from 2D point-based models due to physics engine integration. While 3D collision handling prevents interpenetration, human-to-human interactions require substantial adaptation—agents in 3D may stumble or fall under congestion that would merely slow movement in 2D models. Effective navigation requires detouring behavior rather than waiting or collision. To address these challenges, we introduce personalized SFM coefficient optimization and evasive forces specifically for 3D environments. Our decision process transforms agent $i$'s observations (position, speed, and nearby agents) into desired speed $\tilde{\boldsymbol v}_{i,t+1} \in \mathbb{R}^3$ for the next timestep.

\cg{
\textbf{Base Forces.}
Our implementation enhances the basic social force model \cite{helbing1995social} with two essential components. The driving force guides agents efficiently toward destinations by first employing A* pathfinding \cite{hart1968formal} to generate optimal routes, then calculating acceleration based on the discrepancy between desired and current speeds, smoothed by a relaxation parameter. The complementary repulsive force maintains safe inter-agent distances through exponential repulsion that intensifies as proximity increases, with magnitude controlled by repulsion coefficient, spatial decay, and interaction radius parameters.}

\textbf{Evasive Force. }
 We propose the evasive force to evade the front agents stopping or falling due to collisions. By computing perpendicular directions to the desired path, the agent can detour efficiently. The force is calculated by considering a 45° sector in front of the agent, representing the perceived area. The evasive force $\boldsymbol F_{\text{evasive}}$ is defined as:
 \vspace{-0.7em}
\begin{equation}
    \boldsymbol F_{\text{evasive}}
= A \,\mathrm{sgn}\bigl(\boldsymbol o_{i}\!\cdot\!\boldsymbol p_{i}\bigr)\,\boldsymbol p_{i}\,,
\vspace{-0.7em}
\end{equation}
where A is a binary mask that equals 1 when there is an obstacle in front of the agent and there is available space to the side, and 0 otherwise. $\mathbf{p}_i$ is the perpendicular vector to the desired direction, and $\mathbf{o}_i$ represents the average position of nearby agents. This improvement aligns the SFM with the requirements of 3D simulations.

\cg{
\textbf{Personalized Optimization for SDM Coefficients. } \label{SDM Coefficients}
In crowd evacuation simulations, it \mj{seems not} reasonable to use uniform social force coefficients, as individuals have varying escape abilities based on their attributes. Personalized coefficients better reflect these differences. Moreover, factors, \mj{for example} friction in the physics engine, can cause individuals to move more slowly than the intended speed. To compensate \yl{for these}, we adjust the target speed setting to achieve the desired speed in simulation. Specifically, $v_{\text{real}}$ is the empirically obtained real-world escape speed from literature (see Supplementary), $v_{\text{sim}}$ is the simulated speed, and $v_{\text{setting}}$ is the speed input to the engine. We run agents of different attributes on a straight path in the physics engine and optimize $v_{\text{setting}}$ so that $v_{\text{sim}}$ approximates $v_{\text{real}}$. We categorize agents into five groups, \mj{i.e.} the young, middle-aged, elderly, patients, and disabled\footnote{This classification is solely for scientific purposes, with no intention to offend or make assumptions based on age or disability.}. Through this process, we determine personalized coefficients with an accuracy of 0.005 m/s for use in simulation. The algorithm details are presented in the Supplementary Materials.}

\textbf{Final Decision.}
The final desired speed $\tilde{\boldsymbol v}_{i,t+1}$ is determined by combining the forces:
\vspace{-0.7em}
\begin{equation}
    \tilde{\boldsymbol v}_{i,t+1}
= \boldsymbol v_{i}
+ \Delta t \,\bigl(
  \boldsymbol F_{\text{drive}}
  + \boldsymbol F_{\text{repulsive}}
  + \boldsymbol F_{\text{evasive}}
\bigr),
\vspace{-0.7em}
\end{equation}
where $\Delta t$ is the timestep size, and $\boldsymbol v_{i}$ is the current speed of agent $i$. This framework ensures a smooth, collision-free, and efficient movement toward the destination.


\subsection{Personalized Gait Control} \label{Motor}
\cg{

We propose a Personalized Gait Controller that generates personalized escape gaits by transforming unpersonalized actions to personalized actions based on agent attributes. This section corresponds to the Motor process in \yl{Section} \ref{Framework}. Pacer \cite{rempe2023trace}, a robust physics-based controller, serves as our backbone for path-following. Its policy $\pi_{\mathrm{PACER}}$ tracks 2D trajectories $\tau_{i,t}$, sampled based on the desired speed. Given the state $\mathcal{S}$ (agent’s position, humanoid state, and environment state), the action $\boldsymbol{a}_{i,t}$ is computed by $\pi_{\mathrm{PACER}}$.}

People with different attributes have different gaits. We correspond these gaits to a finite number of personalization styles in the 100style dataset \cite{mason2022real}. This converter is a diffusion-based generative model, taking personalized gait labels, characterless action frames, and randomly sampled Gaussian noise as inputs, and outputs personalized action frames given to the characterless action frames corresponding to the personalized gait labels. This requires the characterless action frames as inputs and the personalized action frames as ground-truths during training should be matched.

\textbf{Gait Frame Matching.} Inspired by \cite{Neumann2002KinesiologyOT}, a gait cycle can be fundamentally divided into 8 events \cite{Neumann2002KinesiologyOT}. We only use 4 events which are \textit{Initial Contact, Mid Stance, Opposite Initial Contact} and \textit{Feet Adjacent}. By calculating the distance between the two ankles, the image can be obtained similar to the sine-cosine function. The four gait events align with two consecutive cycles of the ankle distance waveform, corresponding to the peak, trough, peak, and trough, respectively. To ensure gait connectivity for gait frame matching, we assign \textit{gait values} for all gait frames. We identify the keyframes of four events based on the above pattern and assign gait values of 0, 0.3, 
0.5, and 0.75, while linearly interpolating the non-keyframes between them. We match personalized gait frames (excluding neutral style data in 100STYLE \cite{mason2022real}) and non-personalized gait frames (neutral style data in 100TYLE \cite{mason2022real}) which have the same gait values as data pairs. When forming multiple candidate data frames, we select the two frames with the closest joint angles to match. The algorithm details are presented in the Supplementary Materials. During training, we also perform data augmentation by randomly assigning the same rotation to the root joints of the data pair.

\begin{figure}[h]
\vspace{-1.0em}
  \centering
  \includegraphics[width=\linewidth]{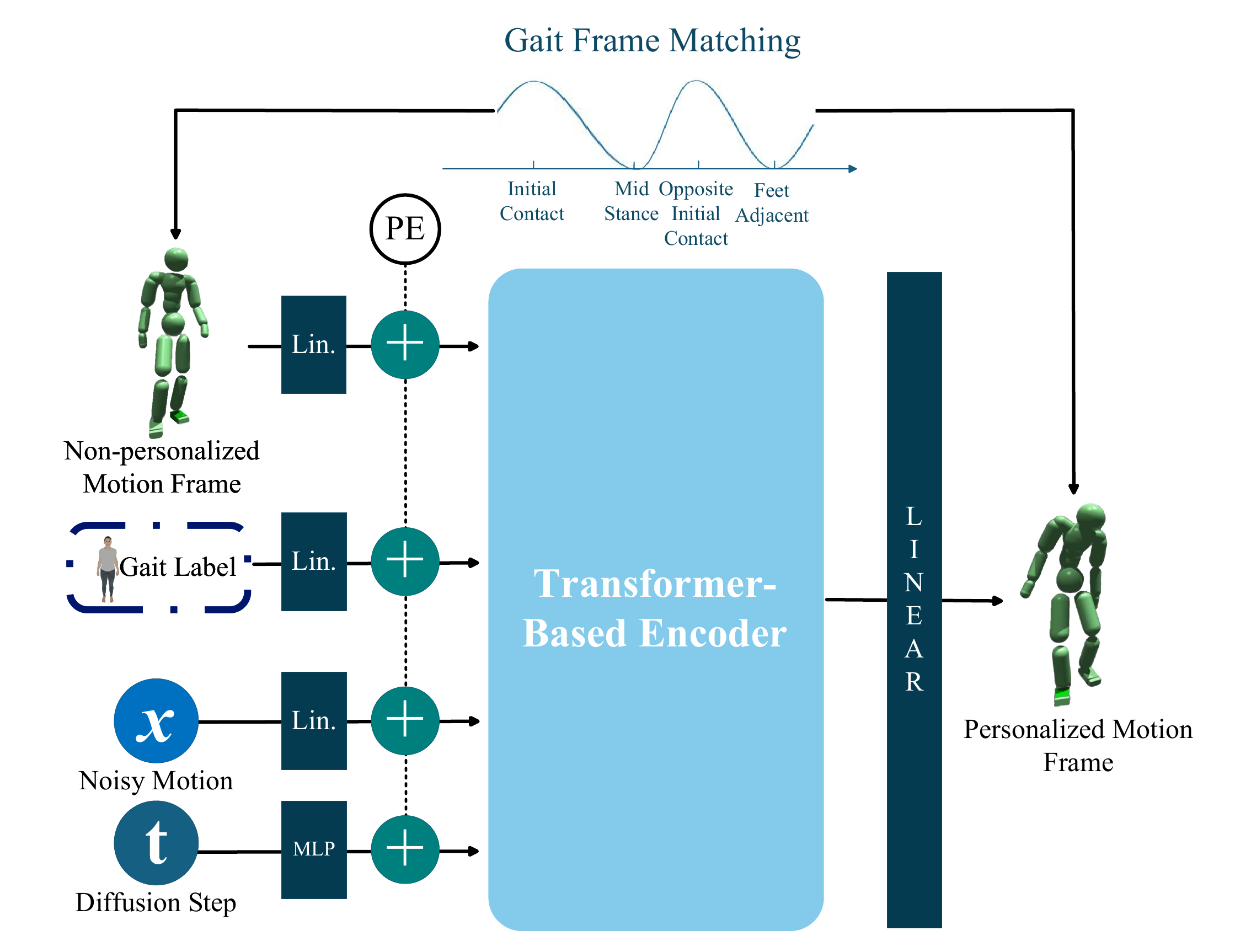}
  \vspace{-2.5em}
  \caption{Personalized Gait Converter.}
  \label{fig:gaitconver}
\vspace{-1.0em}
\end{figure}

\textbf{Network of Personalized Gait Converter.} We use the CAMDM network \cite{chen2024taming} as the backbone to implement the non-personalized action frame  $\boldsymbol{a}_{i,t}$ to personalized action frame $\tilde{\boldsymbol{a}}_{i,t}$, see Figure \ref{fig:gaitconver}. This network is a probability diffusion model. At each denoising step, the model takes as input a noisy motion sample  $\boldsymbol{a}_{i,t}^t$, diffusion step $\mathbf{t}$, along with personalized gait label $c$, then learns to predict the original clean $\tilde{\boldsymbol{a}}_{i,t}^0$ (see Figure \ref{fig:gaitconver}):

\vspace{-1.0em}
\begin{equation}
    \boldsymbol{a}^0=G(\boldsymbol{a}^t,\mathbf{t};c).
\vspace{-0.3em}
\end{equation}
The gait converter only involves single frame action conversion, and we use MSE loss $\mathcal{L}_{\mathrm{samp}}$ and 3D joint position loss $\mathcal{L}_{\mathrm{pos}}$. $\lambda_{\mathrm{samp}}$ and $\lambda_{\mathrm{pos}}$ are hyperparameters that control the weights of $\mathcal{L}_{\mathrm{samp}}$ and $\mathcal{L}_{\mathrm{pos}}$ in the total loss function:

\vspace{-0.5em}

\begin{equation}
    \mathcal{L}=\lambda_{\mathrm{samp}}\mathcal{L}_{\mathrm{samp}}+\lambda_{\mathrm{pos}}\mathcal{L}_{\mathrm{pos}}.
\end{equation}


Lower body movement is affected by terrain, and upper body movement is more flexible \cite{wang2024pacer+}. Hence, in the simulation of the physics engine, we replace the upper body actions of non-personalized action frames $\boldsymbol{a}_{i,t}$ with the upper body actions of personalized action frames $\tilde{\boldsymbol{a}}_{i,t}$. The simulation of actions in the physics engine can be sensed by the sensory process for the decision and motor in the next timestep.

\subsection{Visualization of Part-Level Forces} \label{Visualization}
Each physical agent consists of 24 distinct body parts. Each body part contributes to the overall motion of the physical character. The interactions between these characters, as well as their contact with the environment, generate forces that drive their movements. In evacuation scenarios, assessing the forces acting on each individual is crucial for mitigating potential hazards in crowded situations. We integrate force sensors into each body part of the characters, which are capable of recording the magnitude of \textit{contact forces} exerted at every timestep. To visualize the forces acting on part-level, we design a color-based mapping approach. Specifically, a gradient is employed, with lighter colors indicating lower forces and darker colors representing higher forces. 

\begin{figure}[h]
  \centering
  \includegraphics[width=\linewidth]{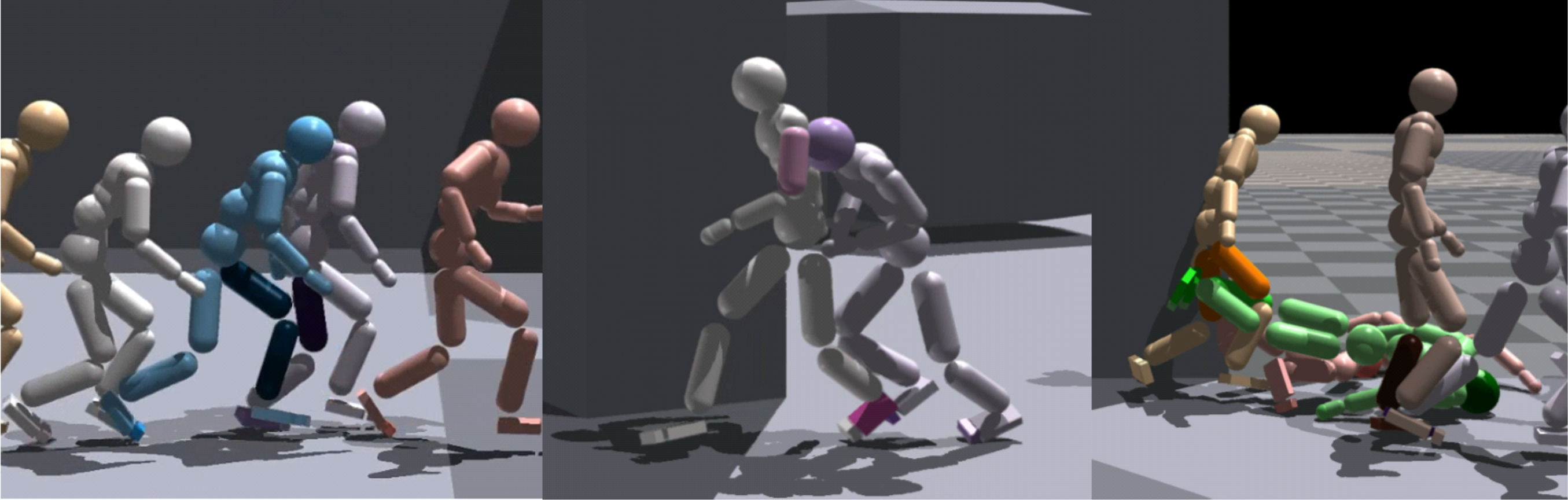}
  \vspace{-2.0em}
  \caption{Visualization of forces under multiple collisions.}
  \label{fig:force_vis}
  \vspace{-1.5em}
\end{figure}

To validate the effectiveness of our force visualization method, we conduct part-level collision simulations under various scenarios. Some examples are shown in Figure \ref{fig:force_vis}.
  
\vspace{-0.5em}


\begin{figure*}[!htbp]

		\centering
		\includegraphics[scale=0.037]{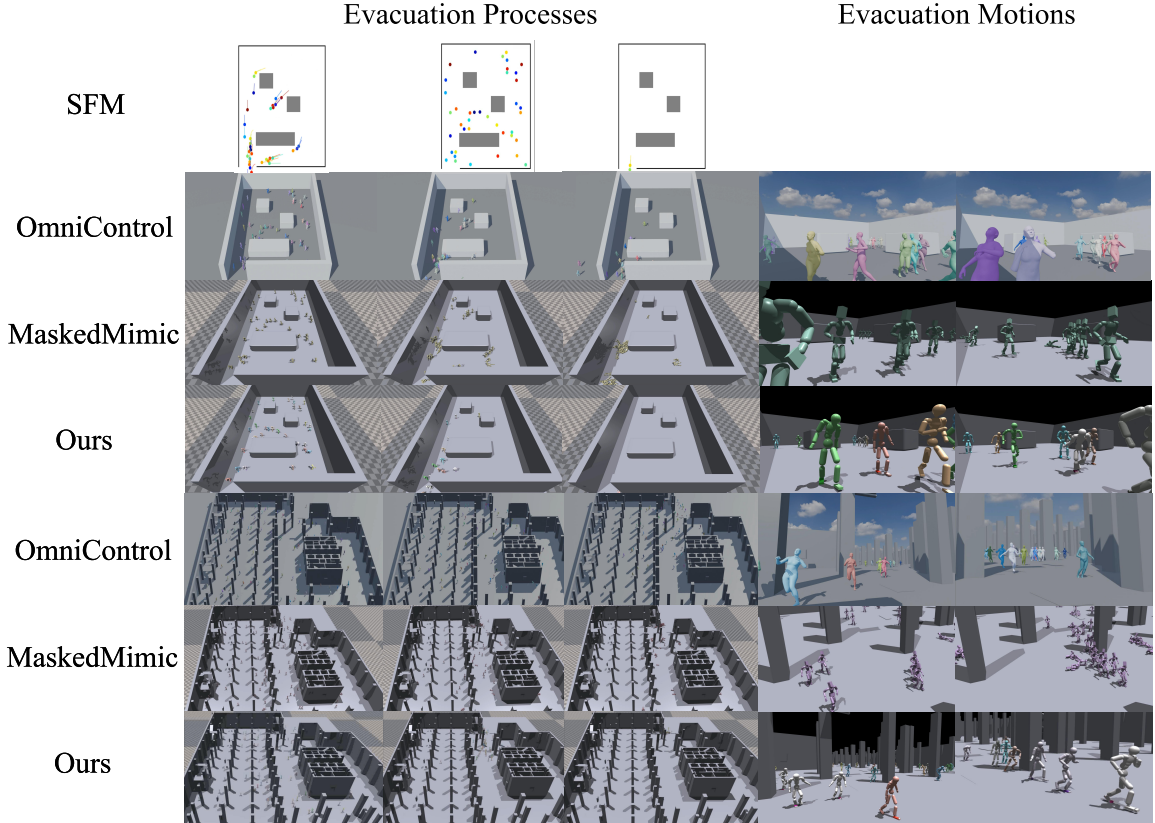}
            \vspace{-1.0em}
		\caption{Qualitative comparison.}
		\label{fig_qualitative}

        \vspace{-1.5em}

	\end{figure*}

\begin{figure*}[!ht]
		\centering
		\includegraphics[scale=0.037]{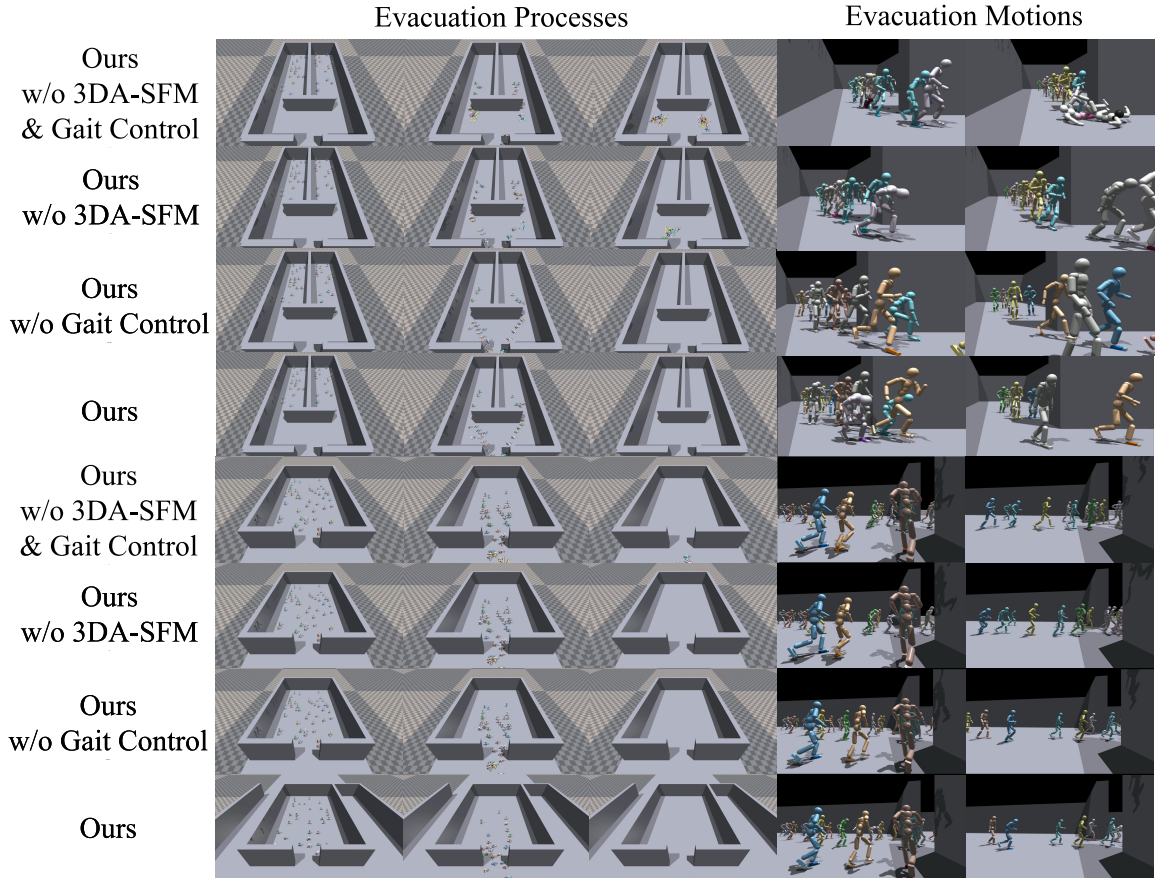}
         \vspace{-1.0em}
		\caption{Ablation study results.}
		\label{fig_ablation}

        \vspace{-2.0em}
        
	\end{figure*}

\section{Experimental Results}

\cg{

\subsection{Comparison Methods}
Since there is no personalized 3D evacuation simulation method,} we compare our approach with three related methods: Social Force Model \cite{helbing1995social,helbing2000simulating}, OmniControl \cite{xie2023omnicontrol} and MaskedMimic \cite{tessler2024maskedmimic}. Among the methods considered, only the Social Force Model \cite{helbing1995social,helbing2000simulating} is originally designed for crowd simulation, while OmniControl \cite{xie2023omnicontrol} is a motion generation method and MaskedMimic \cite{tessler2024maskedmimic} is a character control method. We adapt OmniControl \cite{xie2023omnicontrol} and MaskedMimic \cite{tessler2024maskedmimic} to accomplish the crowd evacuation task as well for a fair comparison. The implementation details are as follows. For all compared methods, we first use A* \cite{hart1968formal} to find waypoints to the exit, and then: (1) \textbf{Social Force Model} \cite{helbing1995social,helbing2000simulating} -- The individual agents of the Social Force Model utilize strategies to reach each waypoint step by step. (2) \textbf{OmniControl} \cite{xie2023omnicontrol} -- We calculate the path and interpolate the path points into dense trajectories, which serve as root joint conditions for generating escape motions. (3) \textbf{MaskedMimic} \cite{tessler2024maskedmimic} -- We compute the escape trajectory, and the path-following task is employed to control the character's escape actions.

 \vspace{-1em}

    \begin{table}[htbp]
    \centering
\small
    \caption{Quantitative comparision.}

 \vspace{-1em}

    \begin{tabular}{@{} l c c @{}}
        \toprule
        \textbf{Method} & \textbf{Average Success Rate} & \textbf{Average Fallen Count} \\
        \midrule
        OmniControl & 0.48 & \textemdash \\
        MaskedMimic & 0.60 & 18.55 \\
        Ours & \textbf{0.84} & \textbf{12.26} \\
        \bottomrule
    \end{tabular}
    \label{tab:table1}
    \vspace{-2.5em}
\end{table}

    \cg{
    

    \subsection{Quantitative Comparison} \label{Qualitative Experiments}

    To validate the performance of our framework in evacuation tasks, we compare it with OmniControl and MaskedMimic across four scenes. We evaluate the average escape success rate and the average number of falls for 50 agents over 1000 frames per scene, with each scene run 10 times. Due to the OmniControl not having physical realism, it does not produce falls, so the fallen Count is not calculated. As shown in Table \ref{tab:table1}, our method outperforms OmniControl and  MaskedMimic in all scenes, with higher success rates and fewer falls. The OmniControl has insufficient controllability for long-distance movements, resulting in agents deviating from trajectories and stopping prematurely. Compared to MaskMinic, in our method, agents demonstrate better pathfinding, mutual avoidance, and balance abilities.
}

 \vspace{-0.5em}
 
    \subsection{Qualitative Comparison} \label{Qualitative Experiments}

To validate the superiority of our proposed framework, we use the above-mentioned three methods for comparison. Figure \ref{fig_qualitative} shows the evacuation processes and evacuation motions executed by the three comparison models and our framework on two classic evacuation scenarios and one large-scale scenario. The Social Force Model is a 2D approach that can only depict the positional changes of people during evacuation, and it cannot represent 3D human movement. It assumes that individuals will always reach their desired position when making decisions, which make the simulation distorted. OmniControl fails to correctly generate long-distance trajectory-constrained actions, leading to motion distortion, trajectory confusion, and premature stopping. MaskedMimic, due to the lack of a collision avoidance mechanism, may cause collisions during linear movement, leading to crowd accumulation. Furthermore, the actions of all agents tend to be similar, failing to reflect individual variations. There is also no noticeable distinction in speed personalization among any of the three methods. Our approach achieves more rational evacuation processes and simulates personalized evacuation motions tailored to individuals with different attributes. More details can be found in demo video and supplementary document.

We also conduct a user study to evaluate our method, along with OmniControl and MaskedMimic, in terms of the plausibility of the evacuation process, the realism and personalization of evacuation motions, and the effectiveness of visualization part-level forces. For more details, please refer to the supplementary document.

\vspace{-1.0em}

\subsection{Ablation Study} \label{ablation}
\vspace{-0.5em}
\textbf{Qualitative comparison.} In qualitative comparison, we compare the pipeline of our framework after removing the 3D-adaptive SFM Decision Mechanism, the Personalizing Gait Control Motor, and both components together, which degrades into the PACER\cite{rempe2023trace} baseline. For all methods, we first find waypoints. Figure \ref{fig_ablation} shows that, when it comes to excluding 3D adaptive SFM while keeping personalized gait control, the generated the agents fail to avoid each other, leading to collisions and crowd accumulation. As a result, some agents are unable to escape, and their escape speeds are nearly identical. This indicates that the 3DA-SFM Decision Mechanism effectively prevents potential collisions as much as possible and offers more rational escape processes tailored to individual attributes. Moreover, although removing the Personalized Gait Control Module allows most agents to fully escape, their gaits and movements become similar. Personalized Gait Control Motor helps to generate personalized motions for agents based on their attributes. In contrast, the complete pipeline can dynamically adjust escape routes based on evacuation speed tailored to each agent's human attributes, minimizing collisions and ensuring personalized escape gaits that reflect individual characteristics. More details can be found in the demo video and supplementary document.

\vspace{-1.0em}

\begin{figure}[!htbp]
  \centering
  \includegraphics[width=0.8\linewidth]{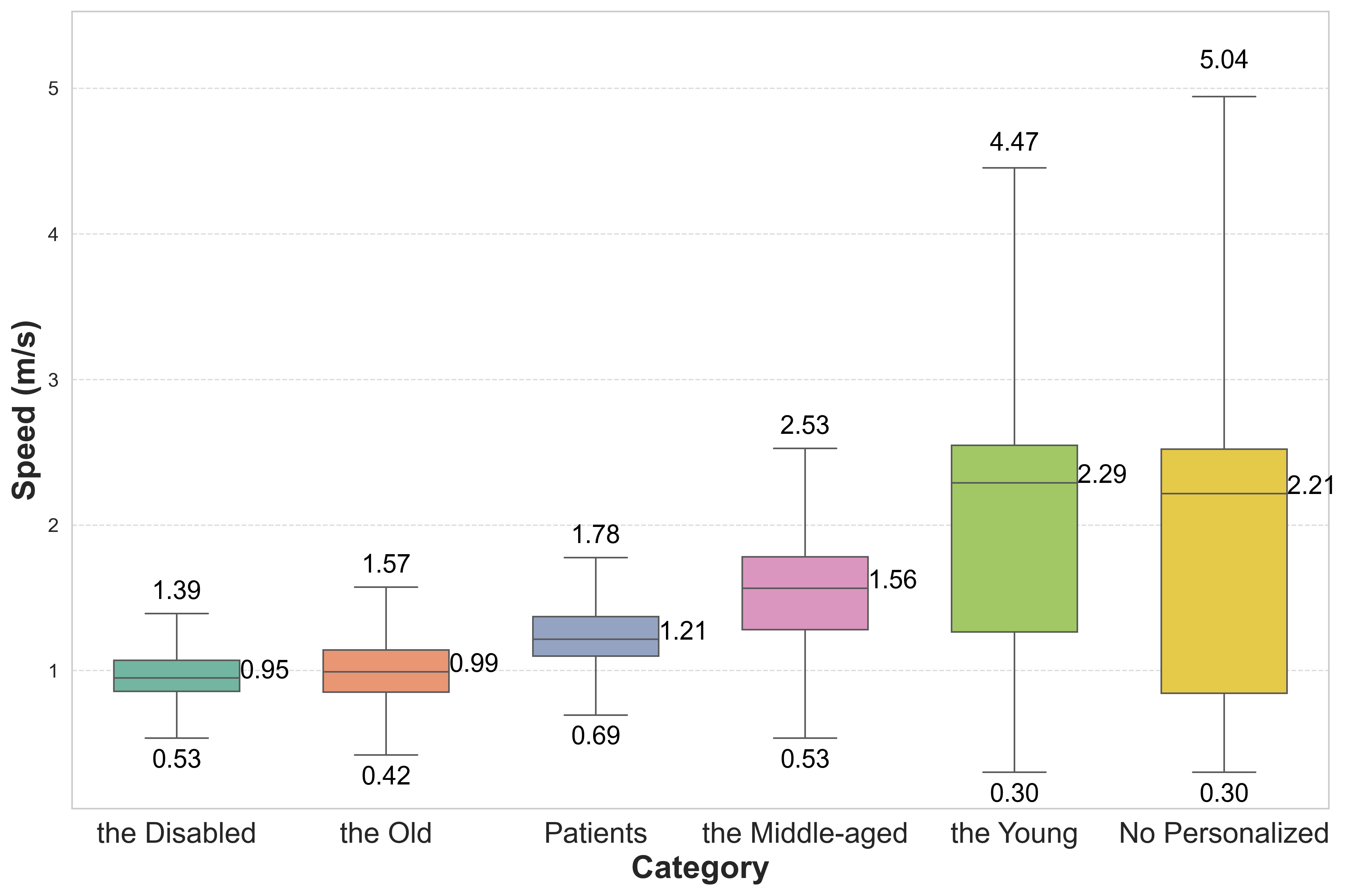}
  \vspace{-1em}
  \caption{Boxplot of speed distributions across categories.}
  \label{fig:speed_boxplot}
  \vspace{-1em}
\end{figure}
\vspace{-0.5em}

\textbf{Speed Diversity.} We conducted experiments across six scenarios, including four small-scale scenarios with 50 agents each and two large-scale scenarios with 100 agents. Agents were divided into five personalized groups—\textit{the young}, \textit{the middle-aged}, \textit{the old}, \textit{patients}, and \textit{the disabled}, as well as a \textit{non-personalized} group. Each category reflects distinct attributes such as age and mobility style.

Figure~\ref{fig:speed_boxplot} illustrates the speed distributions across all categories. The results reveal distinct patterns aligned with mobility attributes. For instance, \textit{the young} achieve the highest median speeds and greater variability, reflecting their agility. In contrast, \textit{the old} and \textit{patients} exhibit slower speeds with narrower ranges, corresponding to their reduced mobility. Similarly, \textit{the disabled} show limited speeds due to physical constraints, while the \textit{non-personalized} group displays average speed distributions.

\vspace{-2.0em}
\begin{figure}[!htbp]
  \centering
  \includegraphics[width=\linewidth]{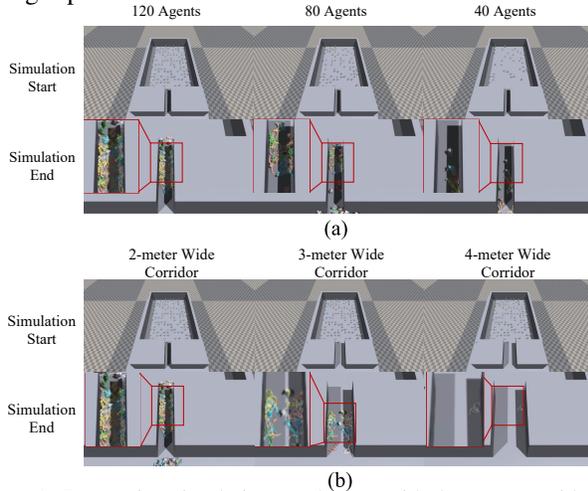}
  \vspace{-2.5em}
  \caption{Evacuation simulation results: (a) with the same corridor width but different evacuation densities; (b) with the same evacuation density but different corridor widths.}
  \label{fig:illu}
  \vspace{-1.5em}
\end{figure}

\subsection{Illustrative Experiment} \label{illustrative}

Our framework can reflect many real-life evacuation in various scenarios. This is more helpful and meaningful  for evacuation analysis.

\textbf{Analysis of Stampede Incidents.} Stampede incidents often occur in bottleneck areas. We analyze the impact of evacuation density and corridor width on stampede occurrences at bottleneck locations. In a 2-meter-wide corridor, we randomly place 120, 80, and 40 agents to simulate evacuation scenarios, and the results are shown in Figure \ref{fig:illu} (a). Under the same corridor width, the greater the number of evacuees, the more severe the stampede. With 120 agents in the same scenario, we test corridor widths of 2 meters, 3 meters, and 4 meters, with the results displayed in Figure \ref{fig:illu} (b). Our results indicate that, for the same evacuation density, narrower corridors lead to more severe trampling.

Analysis of stampede incidents shows they occur when overcrowding leads to falls—from unstable movement—with victims unable to recover, then being stepped on or crushed. Bottlenecks worsen this: large crowds pushing through narrow spaces heighten trampling risks, posing severe hazards. See the demo video for full simulation details.

\begin{figure}[!htbp]
  \centering
  \includegraphics[width=\linewidth]{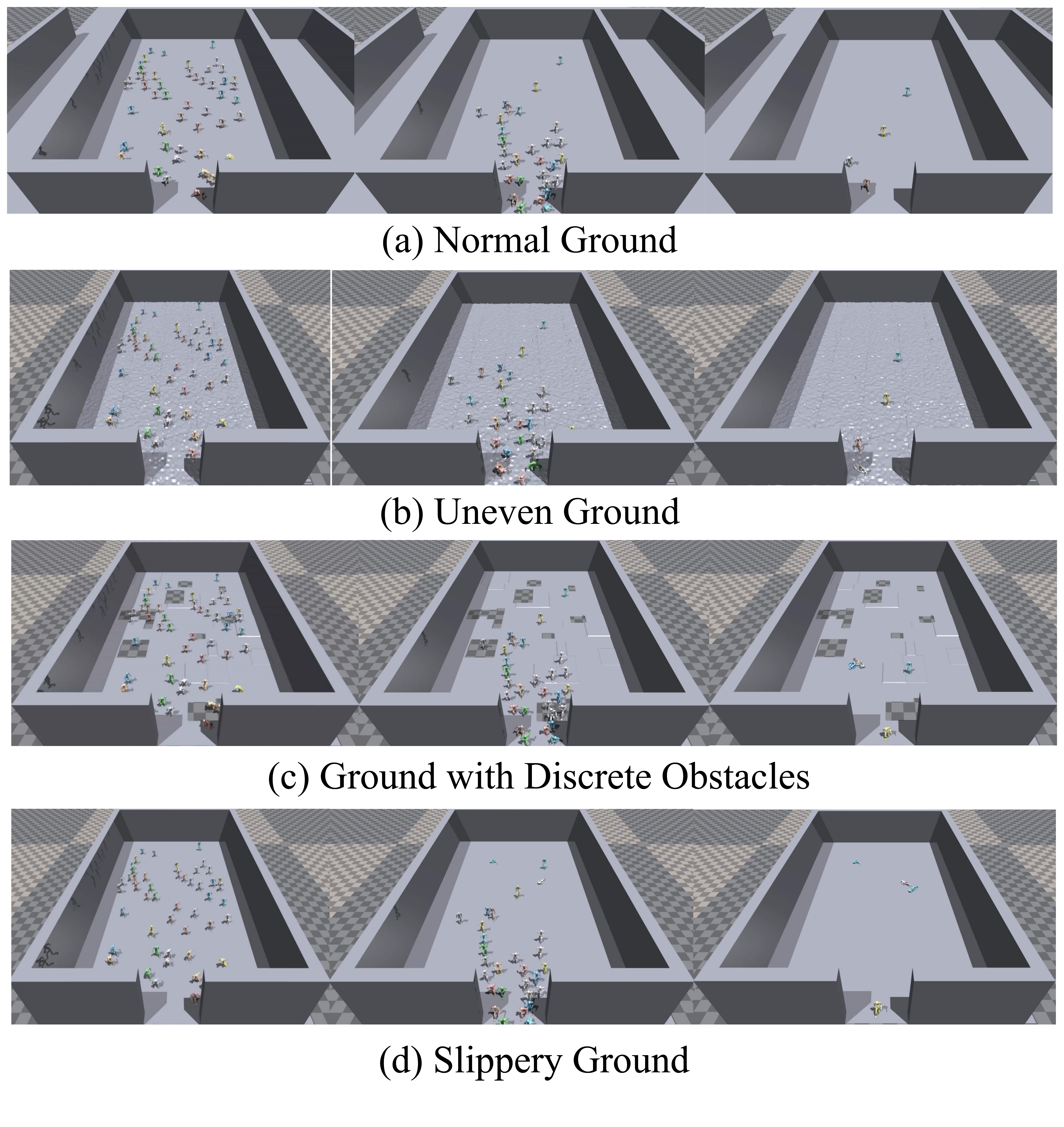}
  \vspace{-3em}
  \caption{Evacuation simulations in various terrains.}
  \label{fig:variousterrain}
  \vspace{-1em}
\end{figure}

\textbf{Analysis of the impact of terrain on evacuation.} Our framework can simulate the impact of various terrains on crowd evacuation. In the same scene, we incorporate normal ground, uneven ground, ground with discrete obstacles, and slippery ground (with reduced friction), and test 50 individuals starting from the same initial position to escape from the same room. Results are presented in Figure \ref{fig:variousterrain}.

Figure \ref{fig:variousterrain} illustrates the different escape processes of individuals within the same scene, using the same number of participants and identical initial positions, but navigating various types of terrain. This demonstrates that terrain factors signifcantly influence motor control and further impact perception and decision-making in subsequent steps, Additionally, uneven and slippery surfaces are more likely to cause individuals to fall.

\vspace{-1.0em}
\vspace{-1.5em}
\section{Conclusions}
\vspace{-0.5em}

We propose a crowd evacuation simulation framework that mimics the sensory-decision-motor flow of a brain-like intelligence, enabling personalized simulations at the individual level. Additionally, we design part-level force visualization, which enhances evacuation analysis. Our framework is capable of simulating the personalized and diverse evacuation dynamics of individuals with varying attributes throughout the evacuation process. The escape motions produced by our method outperform existing methods. Through this framework, we can also validate many common phenomena observed during evacuation scenarios, offering new insights into crowd evacuation and public safety.
\vspace{-1.2em}
\section*{Acknowledgements}
\vspace{-1.0em}
This work was supported in part by the National Key R\&D
Program of China (2023YFC3082100), the National Natural
Science Foundation of China (62171317), the Science Fund for Distinguished Young Scholars of Tianjin
(22JCJQJC00040), and the Engineering and Physical Sciences Research Council (No. EP/Y028805/1).


{
    \small
    \bibliographystyle{ieeenat_fullname}
    \bibliography{main}
}

\end{document}